\pdfoutput=1

\documentclass[11pt]{article}

\usepackage[]{acl}

\usepackage{times}
\usepackage{subfigure}
\usepackage{caption}
\usepackage{subcaption}
\usepackage{graphicx}
\usepackage{latexsym}
\usepackage{amsmath}
\usepackage{algorithm}

\usepackage{amssymb}
\usepackage{adjustbox}
\usepackage{multirow}
\usepackage{soul}
\usepackage{color, xcolor}
\usepackage{makecell}
\usepackage{pifont}
\usepackage{newtxtext,newtxmath}
\DeclareSymbolFont{extraup}{U}{zavm}{m}{n}
\DeclareMathSymbol{\varheart}{\mathalpha}{extraup}{86}
\DeclareMathSymbol{\vardiamond}{\mathalpha}{extraup}{87}

\usepackage[noend]{algpseudocode}

\usepackage[T1]{fontenc}

\usepackage[utf8]{inputenc}

\usepackage{microtype}

\usepackage{inconsolata}

\title{DynClean: Training Dynamics-based Label Cleaning for Distantly-Supervised Named Entity Recognition}

\author{\textbf{Qi Zhang}$^1$ \quad
        \textbf{Huitong Pan}$^1$ \quad 
        \textbf{Zhijia Chen}$^1$ \quad \\
        \textbf{Longin Jan Latecki}$^1$ \quad
        \textbf{Cornelia Caragea}$^2$ \quad
        \textbf{Eduard Dragut}$^1$\\
  $^1$Temple University\quad $^2$University of Illinois Chicago\\ 
  {\tt \{qi.zhang, latecki, edragut\}@temple.edu,  cornelia@uic.edu} \\
  }

\begin{document}
\maketitle
\begin{abstract}
Distantly Supervised Named Entity Recognition (DS-NER) has attracted attention due to its scalability and ability to automatically generate labeled data.
However, distant annotation introduces many mislabeled instances,
limiting its performance. There are two approaches to cope with such instances. One is 
to develop intricate models to learn from the noisy labels. 
Another is to clean the labeled data as much as possible, 
thus increasing the quality of distant labels. The latter approach has received little attention for NER.
In this paper, we propose a training dynamics-based label cleaning approach, which leverages the behavior of a model during training 
to characterize the distantly annotated samples.
We introduce an automatic threshold estimation strategy to locate the errors in distant labels.
Extensive experimental results demonstrate that: 
\ding{182} models trained on our cleaned DS-NER datasets, which were refined by directly removing identified erroneous annotations, achieve significant improvements in F1-score, ranging from 3.19\% to 8.95\%; and \ding{183} our method outperforms state-of-the-art 
DS-NER approaches on four benchmark datasets.
\end{abstract}

\section{Introduction}

Named entity recognition (NER), which aims to identify spans in text that refer to pre-defined entity types such as person, location, or organization names, has wide downstream applications such as question answering \citep{ijcai2021p611}, information retrieval \citep{choudhary-etal-2023-iitd}, and knowledge graph construction \citep{9416312, zhang-etal-2024-scier}.
However, NER performance often relies on human annotated high-quality training data, which is time-consuming and expensive, making the development of robust NER models for real-world applications hard.
Distantly Supervised Named Entity Recognition (DS-NER) \citep{ren2015clustype, fries2017swellshark, shang-etal-2018-learning,KDDDSNER, xu-etal-2023-sampling,DMDD,wu-etal-2023-mproto,BhowmickDM23b, pan-etal-2024-scidmt} has been proposed to automatically generate labeled data for training NER models.
In general, DS-NER leverages existing resources such as knowledge bases and rule-based string matching methods
to automatically produce the training data, which is generally referred to as \textit{weakly annotated data} \citep{zhang2021denoising}. 
Since unlabeled data is easy to collect, DS-NER can significantly reduce the annotation efforts and obtain large-scale datasets.
The weakly annotated data, however, suffers from the introduction of noisy labels.
There are two typical issues in DS-NER: (1) False Negative:  when an entity is not detected due to the limited coverage of knowledge bases
and (2) False Positive:  when  
annotated spans of text is incorrectly detected 
as entities or assigned incorrect entity types.

\begin{figure}[t]
\centering
\includegraphics[width=\linewidth]{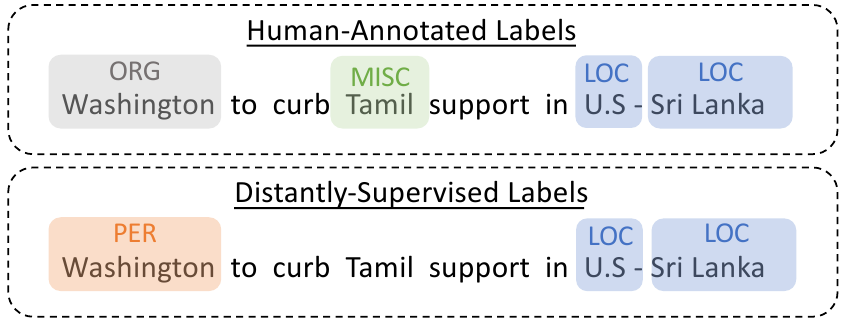}
\vspace{-10pt}
\caption{A typical distantly-supervised annotation can be subject to two types of error: (1) False Positive: An entity is recognized to incorrectly type, e.g., ``Washington'' and (2) False Negative: An entity is recognized as non-entity, e.g.,``Tamil''.}
\label{fig:ds-example}
\vspace{-15pt}
\end{figure}

For example, in Figure \ref{fig:ds-example} 
the organization entity ``Washington'' was incorrectly annotated as a person type,  
since the rule-based string matching cannot distinguish between two entity types (person and organization in this case) with the same surface form; ``Tamil'' is not recognized as an entity because it is not included in the knowledge base used.

Models trained on such DS-NER dataset can easily overfit to the noisy labels, leading to poor performance \citep{tanzer-etal-2022-memorisation}.
Many approaches have been proposed to alleviate these limitations.
Negative sampling \citep{li2021empirical,li-etal-2022-rethinking,xu-etal-2023-sampling} 
and positive unlabeled (PU) learning \citep{peng-etal-2019-distantly,zhou-etal-2022-distantly,wu-etal-2023-mproto}
have been proposed to handle the false negative annotation issues in DS-NER.
But these works overlook
false positives, which limit their performance.
Another line of work employs self-training \citep{meng-etal-2021-distantly,zhang-etal-2021-improving-distantly,KDDDSNER} 
by utilizing soft labels generated by a teacher model. 
They require a high-quality teacher model to produce reliable soft labels for the student model, often resulting in the inclusion of additional modules. 
A common challenge with self-training methods is their complex architectures, which require the training of multiple models across several iterations.
Despite those efforts, the research on DS-NER has yet to thoroughly investigate the improvement of the quality of distantly annotated labels.

In this paper, we propose the DynClean, a training dynamics-based label cleaning approach for DS-NER. 
Unlike most existing methods that focus on learning from noisy labels, \textit{our proposed approach aims to improve the quality of data generated by distant supervision annotation}.
In DynClean, we leverage the training dynamics (i.e., the behavior of a model as training progresses) to reveal the characteristics of each data sample in DS-NER datasets.
To differentiate between clean and mislabeled samples using the obtained data characteristics, we introduce an automatic threshold estimation strategy.
We show that DynClean can effectively locate 
many false positive and negative samples. 
Using its output, we proceed to filter out the identified mislabeled distant samples and  obtain ``mostly clean'' data for training.
We validated our proposed approach across different NER models on four popular DS-NER datasets and achieved a consistent performance boost, with improvements in the F1 score ranging from 3.18\% to 8.95\%.
These results empirically demonstrate that enhancing the quality of distantly annotated data improves the performance of DS-NER tasks.
Furthermore, models trained on four DS-NER datasets cleaned with DynClean outperform current state-of-the-art methods, achieving up to 4.53\% F1 score improvements despite using fewer samples in training \footnote{Code and Dataset are publicly available: \url{https://github.com/maxzqq/DynClean}}.

\section{Related Work} 
\textbf{Distantly-Supervised NER.} \quad 
Due to its ability to reduce annotation costs, DS-NER has gained significant popularity in domain-specific NER or IE \cite{AlshehriKRDO23,ChenMD22,BhowmickDM23a,ritter2013modeling, pan2024flowlearn, dash-ws-2024-data}, where high-quality labeled data is often scarce and expensive to obtain.
Several Negative sampling \citep{li2021empirical,li-etal-2022-rethinking,xu-etal-2023-sampling} methods have been proposed to reduce the negative impact of unlabeled entities in training data.
Other works leverage the positive unlabeled (PU) learning and include potentially noisy positive samples along with unlabeled data \citep{peng-etal-2019-distantly,zhou-etal-2022-distantly}.
Another line of works is based on the self-training strategy \citep{jie-etal-2019-better,KDDDSNER,zhang-etal-2021-improving-distantly-supervised,ma-etal-2023-noise,meng-etal-2021-distantly, BhowmickDM22}.
In general, they design a teacher-student network to iteratively refine the entity labels and reduce both false positive and false negative samples.
\citet{ying-etal-2022-label} proposed CReDEL to leverage contrastive learning to learn the refinement knowledge of distant annotation.
\citet{wu-etal-2023-mproto} design the MProto to model the token-prototype assignment problem as an optimal transport problem to handle intra-class variance issues.
In our work, we introduce a training dynamics-based method that denoises DS-NER datasets automatically without requiring additional components.

As discussed in \citep{zhou-etal-2022-distantly,xu-etal-2023-sampling}, self-training based methods are post-processing steps. 
Thus, 
self-training \citep{KDDDSNER, zhang-etal-2021-improving-distantly-supervised, meng-etal-2021-distantly,ma-etal-2023-noise,si-etal-2023-santa, wang2024debiased}
approaches are orthogonal to data cleaning approaches (like ours), which are pre-processing steps; we do not discuss them further in this paper.

\noindent\textbf{Training Dynamics.} \quad 
Training dynamics are defined as statistics calculated of the model behavior (e.g., the output logit values) on each sample across the training process.
It can be used to evaluate the characteristics and quality of each individual sample within a dataset.
The frequently used metrics include Confidence and Variability proposed by \citet{swayamdipta-etal-2020-dataset}, and the Area Under Margin (AUM) proposed by \citet{pleiss2020identifying}.
For each sample, Confidence and Variability are the mean and standard deviation of the gold label probabilities over the training epochs, respectively.
AUM ﬁnds the difference in logit value of the assigned class (gold label) and the highest logit value among the non-assigned classes averaged over training epochs for every sample.
These metrics capture the data properties and can be used to evaluate the data quality.
Most works obtained the data characteristics from the training dynamics and combined it with different techniques, such as data augmentation \citep{park-caragea-2022-data, cosma-etal-2024-hard}, curriculum learning \citep{sar-shalom-schwartz-2023-curating, poesina-etal-2024-novel} and Semi-Supervised Learning \cite{sadat-caragea-2022-learning, sadat-caragea-2024-co}.
In our work, we adapt the training dynamics metrics to diagnose the DS-NER datasets and identify the noisy instances.

\section{Method}
We introduce our proposed method, DynClean. 
DynClean leverages a model's behavior on individual samples in the DS-NER across training epochs (i.e., the training dynamics) to create characteristics of data samples.
Then, using the proposed automatic threshold estimation strategy, we distinguish clean and mislabeled data within the DS-NER dataset.
In this section, we first give a brief overview of the span-based NER model, 
followed by a detailed presentation of the proposed DynClean.

\subsection{Background} \label{Method:NER_model}
\textbf{Span-based NER Model.}\quad
Following previous works \citep{li-etal-2022-rethinking, si-etal-2023-santa}, we employ the span-based NER architecture. Given a sentence with $n$ tokens, $[t_1, t_2, ..., t_n]$, we take a pre-trained language model as the encoder to get the contextualized representation:
\begin{equation}
    [ \mathbf{h}_1, \mathbf{h}_2, ..., \mathbf{h}_n] = \textrm{Encoder}([t_1, t_2, ..., t_n])
\end{equation}
where $\mathbf{h}_i$ is the representation for token $t_i$. We derive the representation for one span sample $x$ which starts at position $i$ and ends at position $j$ as:
\begin{equation}
    \mathbf{x} = \mathbf{h}_i \oplus \mathbf{h}_j \oplus \mathbf{D}(j-i)
\end{equation}
where $\oplus$ is the concatenation operation, $\mathbf{D}(j-i)$ is a learnable embedding to encode the span length feature, and the span length is limited to $0\leq j-i \leq L$. 
A feed-forward neural network (FFNN) is used to obtain the logits $\mathbf{z}$, and $\textrm{softmax}$ computes the probability distribution of label $y$:
\begin{equation} \label{eq:logits}
    \mathbf{z} = \textrm{FFNN}(\mathbf{x})
\end{equation}
\begin{equation} \label{eq:prob}
    P(y|\mathbf{x}) = \textrm{softmax}(\mathbf{z})
\end{equation}

\noindent\textbf{Training.}\quad Cross-entropy loss function is calculated on all spans during training: 
\begin{equation} \label{eq:loss}
    \mathcal{L} = -\sum_{\mathbf{x}\in \mathbf{X}} \log P(y^*|\mathbf{x})
\end{equation}
where $y^*$ represents the target label.

\noindent\textbf{Negative Sampling.}\quad Previous studies \citep{li-etal-2022-rethinking,li2021empirical,xu-etal-2023-sampling} have demonstrated that negative sampling can effectively mitigate the impact of false negatives in DS-NER.
Motivated by the \citep{xu-etal-2023-sampling}, we consider only utilizing the top-$N_r$ negative samples (TopNeg), which exhibit high similarities with all positive samples, for training the DS-NER model.
TopNeg reduces the number of false negative samples involved in training, thereby enhancing DS-NER performance.
Specifically, for each batch of training data, we obtain the span sets of positive samples $\mathbf{X}^{\textrm{pos}} = \{..., \mathbf{x}^{\textrm{pos}}, ...\}$, and negative samples $\mathbf{X}^{\textrm{neg}} = \{..., \mathbf{x}^{\textrm{neg}}, ...\}$, where $\mathbf{X} = \mathbf{X}^{\textrm{pos}} \cup \mathbf{X}^{\textrm{neg}}$. 
Subsequently, for each negative sample $\mathbf{x}^{\textrm{neg}} \in \mathbf{X}^{\textrm{neg}}$, we compute the similarity score $\phi$:
\begin{equation} \label{eq:sim}
    \phi(\mathbf{x}^{\textrm{neg}}, \mathbf{X}^{\textrm{pos}}) = \frac{1}{M}\sum_{\mathbf{x}^{\textrm{pos}}\in \mathbf{X}^{\textrm{pos}}} \frac{\mathbf{x}^{\textrm{neg}}}{\left \| \mathbf{x}^{\textrm{neg}} \right \| } \cdot \frac{\mathbf{x}^{\textrm{pos}}}{\left \| \mathbf{x}^{\textrm{pos}} \right \| }
\end{equation}
where $M$ denotes the number of positive samples in one batch of data.

When training the model with TopNeg, the negative samples in each batch are ranked ascending according to their similarity scores, as defined in Equation \ref{eq:sim}. 
Subsequently, the top-$N_r$ negative samples along with all positive samples, are selected for loss calculation.  Equation \ref{eq:loss} is modified as:
\begin{equation}
\begin{aligned}
    \mathcal{L} = -\sum_{\mathbf{x}^{\textrm{pos}}\in \mathbf{X}^{\textrm{pos}}} \log P(y^*|\mathbf{x}^{\textrm{pos}}) \\
    -\sum_{\tilde{\mathbf{x}}^{\textrm{neg}}\in \tilde{\mathbf{X}}^{\textrm{neg}}} \log P(y^*|\tilde{\mathbf{x}}^{\textrm{neg}})
\end{aligned}
\end{equation}
and $\tilde{\mathbf{X}}^{\textrm{neg}}$ is the set of the selected negative samples.

\subsection{Proposed Approach: DynClean} \label{sec:proposed_method} \label{sec:denoising_method}

We detail our proposed method, DynClean, which differentiates clean and mislabeled data based on the characteristics of each sample, derived from the training dynamics.
We first define the training dynamics and corresponding metrics; next we show how we use them to characterize each data sample.
We then present our threshold estimation strategy and procedure DynClean of cleaning distant labels.

\noindent\textbf{Defining training dynamics.}\quad
Training dynamics can be defined as any model behavior during the training, such as the area under the margin (AUM) between logit values of the target label and the other largest label \citep{pleiss2020identifying} or the mean of predicted probabilities (Confidence) of target label \citep{swayamdipta-etal-2020-dataset}.
This section focuses on AUM in our primary experiments, with a discussion on Confidence in Section \ref{sec:confidence}.

Given a sample $\mathbf{x}$ with assigned label $y^*$ (potentially erroneous), we compute $\textrm{AUM}(\mathbf{x}, y^*)$ as the area under the margin averaged across all training epochs $E$.
The margin at epoch $e$ is defined as:
\begin{equation} \label{eq:margin}
    M^{e}(\mathbf{x}, y^*) = \mathbf{\mathbf{z}}^{e}_{y^*}(\mathbf{x}) - \textrm{max}_{k\neq y^*}\mathbf{\mathbf{z}}_{k}^{e}(\mathbf{x})
\end{equation}
where $M^{e}(\mathbf{x}, y^*)$ is the margin of sample $\mathbf{x}$ with label $y^*$, $\mathbf{z}^{e}_{y^*}(\mathbf{x})$ is the logit corresponding to the label $y^*$, and $\textrm{max}_{k\neq y^*}\mathbf{z}_{k}^{e}(\mathbf{x})$ is the largest non-assigned logit with label $k$ not equal to $y^*$. 
The margin measures the difference between an assigned label and the model's predictions at each epoch $e$.
The $\textrm{AUM}(\mathbf{x},y^*)$ across $E$ epochs is computed with:
\begin{equation} \label{eq:AUM}
    \textrm{AUM}(\mathbf{x},y^*) = \frac{1}{E}\sum_{e=1}^{E}M^{e}(\mathbf{x}, y^*)
\end{equation}
Precisely, AUM considers each logit value and measures how much the assigned label logit value differs from the other largest logit value.
A low AUM signifies that the assigned label is likely incorrect, while a larger margin indicates the assigned label is likely correct.
Consequently, AUM can be utilized to distinguish mislabeled samples.

\noindent\textbf{Threshold Estimation.}\quad
A threshold value is required to separate mislabeled from clean samples. 
However, given the varying noise distribution across different datasets, determining this threshold is a costly hyper-parameter tuning procedure.
Thus, we construct threshold samples to simulate mislabeled data, and train the model to compute the training dynamics metric values for threshold samples.
Data with similar or worse metric values than threshold samples can be assumed as mislabeled.
We select a subset of both positive and negative samples from the training data and reassign their labels to a non-existent class, thereby constructing threshold samples.
Assuming we have $N_{\textrm{p}}$ positive samples (i.e., entity spans that belong to $c$ entity types). 
We use stratified sampling (based on the distribution of the original entity types) to randomly select $N_{\textrm{p}}/(c+1)$ samples as positive threshold samples. 
Then we assign a fake label $c+1$ to these positive threshold samples. 
As for negative threshold samples, we sample the same number ($N_{\textrm{p}}/(c+1)$) of negative samples and assign the fake label $c+1$ as well.
A model is trained on the constructed datasets with threshold samples to compute the metric values at the end of training.
Positive and negative threshold samples are ranked separately, employing the threshold samples at the $k_{\textrm{pos}}$th for positives and $k_{\textrm{neg}}$th for negatives percentile to determine the positive threshold $\tau_{pos}$ and negative threshold $\tau_{neg}$.
Constructing threshold samples for both positives and negatives is crucial for identifying unlabeled negatives and mislabeled positives in the distant annotation process in our DS-NER case.

\begin{algorithm}[t]
\small
\caption{DynClean}\label{alg:noise_filter}
\begin{algorithmic}[1]
\Require Distantly annotated data $\mathcal{D} = {(x_i, y_i)}_{i=1,...,n}$, model $f$, hyperparameter $k_{\textrm{pos}}$ and $k_{\textrm{neg}}$; 
\State // Threshold Estimation
\State{$T_{\textrm{pos}}\leftarrow$ Stratified sampling $N_{\textrm{p}}/(c+1)$ positives}
\State{$T_{\textrm{neg}}\leftarrow$Random sampling $N_{\textrm{p}}/(c+1)$ negatives}
\State{Train $f$ for $E$ epochs and compute $\textrm{AUM}$ for each sample in $T_{\textrm{pos}}$ and $T_{\textrm{neg}}$ as in Eq. \ref{eq:AUM} }
\State{Rank AUM for $T_{\textrm{pos}}$ and $T_{\textrm{neg}}$ separately}
\State{$\tau_{\textrm{pos}}\leftarrow$ $k_{\textrm{pos}}$th at $T_{\textrm{pos}}$}
\State{$\tau_{\textrm{neg}}\leftarrow$ $k_{\textrm{neg}}$th at $T_{\textrm{neg}}$}
\State // Noisy data cleaning
\State{$\mathcal{D}^{\prime} \leftarrow$ $\varnothing$, re-initialize model $f$}
\State Train $f$ for $E$ epochs and compute $\textrm{AUM}(x_i, y_i)$ for each sample $i$ as in Eq. \ref{eq:AUM}
\For{each $(x_i, y_i) \in \mathcal{D}$}
    \If{$y_i > 0$} \Comment{Positive samples}
        \If{$\textrm{AUM}(x_i, y_i) \geq \tau_{\textrm{pos}}$}
        \State{$\mathcal{D}^{\prime} \leftarrow$ $\mathcal{D}^{\prime} \cup (x_i, y_i)$ }
        \EndIf
    \Else \Comment{Negative samples}
        \If{$\textrm{AUM}(x_i, y_i) \geq \tau_{\textrm{neg}}$}
        \State{$\mathcal{D}^{\prime} \leftarrow$ $\mathcal{D}^{\prime} \cup (x_i, y_i)$ }
        \EndIf
    \EndIf
\EndFor
\State \Return $\mathcal{D}^{\prime}$
\end{algorithmic}
\end{algorithm}

\noindent\textbf{Applying DynClean.}\quad 
Given a DS-NER dataset, we need to utilize a model to train on it in order to gather training dynamics metric values. 
Specifically, we begin with the proposed automatic threshold estimation strategy, which involves constructing threshold samples and collecting their training dynamics to compute the corresponding metric values. 
Subsequently, we collect the training dynamics metric values for all samples in the original DS dataset. Finally, we separately filter the positive and negative samples based on the estimated threshold.
Thus, our method can be applied to various models. 
The details of our DynClean procedure when using AUM are summarized in Algorithm \ref{alg:noise_filter}.
Our method is also applicable to other training dynamic metrics, as presented in Section \ref{sec:confidence}.

\section{Experiments}

\subsection{Datasets}
Our approach is evaluated using four DS-NER datasets: CoNLL03 \citep{tjong-kim-sang-de-meulder-2003-introduction}, WikiGold \citep{balasuriya-etal-2009-named}, WNUT16 \citep{godin-etal-2015-multimedia}, and BC5CDR \citep{wei2016assessing}.
Originally, these datasets were human-annotated, and subsequently, they were re-annotated via distant supervision as reported in \citep{KDDDSNER, zhou-etal-2022-distantly, shang-etal-2018-learning}.
The statistics for the four datasets are presented in Table \ref{tab:dataset_stat} of Appendix \ref{apx:dataset_stat}.
The distantly supervised data are used for training, while the human-annotated development and test sets are utilized for evaluation and hyperparameter tuning, aligning with the general DS-NER setting \cite{KDDDSNER, wu-etal-2023-mproto}.

\subsection{Experimental Setup}
To validate the effectiveness of our proposed method, we first conduct cleaning experiments on four datasets with our DynClean framework using different base models.
Subsequently, we compare the models trained on our cleaned datasets with existing DS-NER studies.
We describe below the base models used in our work.

\noindent\textbf{Base Models.}\quad
We employ span-based NER model mentioned in Section \ref{Method:NER_model}, integrating different encoders and their TopNeg variants, as our base models.
We evaluate BERT \citep{devlin-etal-2019-bert} and RoBERTa \citep{liu2019roberta} as encoders, along with their TopNeg variants, on CoNLL03, Wikigold, and WNUT16.
The four base models are denoted by BERT, BERT-TopNeg, RoBERTa, and RoBERTa-TopNeg.
For BC5CDR, a domain-specific NER task, we exclusively employ BioBERT \citep{lee2020biobert}.
For ease of presentation, in the context of BC5CDR, BERT and BERT-TopNeg are used to represent its base models.

\noindent\textbf{Baselines.}\quad 
We 
compare our method with several existing DS-NER works, including:
\textbf{KB-Matching} employs knowledge bases to annotate the test sets.
\textbf{AutoNER} \citep{shang-etal-2018-learning} modifies the standard CRF to handle noisy labels.
\textbf{BOND} \citep{KDDDSNER} proposes an early stopping approach to prevent model from overﬁtting to noisy labels.
\textbf{bnPU} \citep{peng-etal-2019-distantly} formulates the DS-NER task as a positive unlabelled learning problem and uses the mean absolute error as the noise-robust objective function.
\textbf{Span-NS} \citep{li2021empirical}, \textbf{Span-NS-V} \citep{li-etal-2022-rethinking}
are the negative sampling approaches, which aim to reduce the false negative samples used for training.
\textbf{CReDEL} \citep{ying-etal-2022-label} adopts a distant label refinement model via contrastive learning.
\textbf{Conf-MPU} \citep{zhou-etal-2022-distantly} is a two-stage approach, with the first stage estimating the confidence score of being an entity and the second stage incorporating the confidence score into the positive unlabelled learning framework. 
\textbf{MProto} \citep{wu-etal-2023-mproto} models the token-prototype assignment problem as an optimal transport problem to handle intra-class variance issue. 

We also provide fully supervised methods for comparison, including RoBERTa and BiLSTM-CRF \citep{ma-hovy-2016-end}. We also conduct zero-shot evaluations with large language models (LLMs), including ChatGPT and LLaMA3.1-70B. 

\subsection{Implementation Details}

We tune the $k$th of the threshold samples for threshold estimation, with increments of 5\%, ranging from 80\% to 100\%.
The $k_{\textrm{neg}}$th for computing the threshold in negative threshold samples is set at 90\% across all datasets.
For positive threshold samples, the tuned $k_{\textrm{pos}}$th is set at 100\% for CoNLL03, Wikigold, and WNUT16, and at 90\% for BC5CDR.
A more detailed discussion on the effectiveness of our threshold estimation strategy can be found in the Appendix \ref{apx:thre_estimate_analysis}.
All reported experimental results represent the average of five runs, each with a different random seed.
More experimental details are provided in the Appendix \ref{apx:addional_settings}.

\begin{table*}[htb] 
\small
\centering
\begin{adjustbox}{width=1\textwidth}
\begin{tabular}{lccccccccccccc} 
\hline
\multirow{2}{*}{Models} & \multirow{2}{*}{Data Ver.} & \multicolumn{3}{c}{CoNLL03} & \multicolumn{3}{c}{WikiGold} & \multicolumn{3}{c}{WNUT16} & \multicolumn{3}{c}{BC5CDR}  \\ 
\cline{3-14}
                                 &                                 & P     & R     & F1          & P     & R     & F1           & P     & R     & F1         & P     & R     & F1          \\ 
\hline
\multirow{4}{*}{BERT}            & $\mathcal{D}$                            & \textbf{89.75}                     & 63.08                     & 72.09                     & 56.12                     & 44.37                     & 49.47                     & 58.18                     & 36.22                     & 44.55                     &  \textbf{89.44}                    & 67.57                     & 76.98                      \\
                                 & $\mathcal{D}^{\prime}_1$                         & 88.77 & 65.61 & 75.44       & 58.18 & 46.29 & 51.56        & 57.60 & 38.19 & 45.93      & 87.07 & 71.67 & 78.62       \\
                                 & $\mathcal{D}^{\prime}_2$                     & 89.19 & 67.92 & 77.11       & 58.09 & 47.94 & 52.53        & 58.54 & 37.98 & 46.02      & -     & -     & -           \\
                                 & $\mathcal{D}^{\prime}_3$                  & 86.79 & 77.12 & 81.66       & 47.24 & \textbf{57.91} & 52.02        & 53.25 & \textbf{41.60} & 46.51      & 80.02 & \textbf{83.06} & \textbf{81.50}       \\ 
                                 & $\mathcal{D}^{\prime}_4$                  & 87.29 & \textbf{80.40} & \textbf{83.70}       & \textbf{60.75} & 56.71 & \textbf{58.65}        & \textbf{58.75} & 40.25 & \textbf{47.74}      & -     & -     & -           \\ 
\hline
\multirow{4}{*}{RoBERTa}         & $\mathcal{D}$                            & \textbf{90.16}                     & 65.88                     & 76.13                     & 60.63                     & 44.65                     & 51.42                     & \textbf{62.72}                     & 43.45                     & 51.34                     & -                         & -                         & -                          \\
                                 & $\mathcal{D}^{\prime}_1$                          & 89.99 & 67.76 & 77.31       & 57.92 & 48.30 & 52.65        & 61.85 & 43.65 & 51.18      & -     & -     & -           \\
                                 & $\mathcal{D}^{\prime}_2$                      & 89.51 & 70.10 & 78.61       & 56.47 & 51.12 & 53.66        & 62.33 & 43.86 & 51.44      & -     & -     & -           \\
                                 & $\mathcal{D}^{\prime}_3$                   & 87.96 & 81.47 & 84.59       & 51.24 & \textbf{61.90} & 56.02        & 61.34 & 47.44 & 53.50      & -     & -     & -           \\
                                 & $\mathcal{D}^{\prime}_4$                   & 88.41 & \textbf{82.77} & \textbf{85.50}       & \textbf{60.94} & 59.31 & \textbf{60.09}        & 58.22 & \textbf{51.35} & \textbf{54.52}      & -     & -     & -           \\ 
\hline
\multirow{4}{*}{BERT-TopNeg}     & $\mathcal{D}$                          & 79.42                     & 79.77                     & 79.58                     & 52.99                     & 47.83                     & 50.09                     & 49.05                     & \textbf{42.96}                     & 45.80                     & 79.22                     & 81.35                     & 80.27                      \\
                                 & $\mathcal{D}^{\prime}_1$                         & 82.09 & 77.98 & 79.98       & 57.59 & 47.45 & 51.98        & \textbf{58.56} & 37.92 & 46.03      & \textbf{82.79} & 76.52 & 79.53       \\
                                 & $\mathcal{D}^{\prime}_2$                     & 85.01 & 77.08 & 80.84       & 58.17 & 47.72 & 52.43        & 58.35 & 38.96 & 46.71      & -     & -     & -           \\
                                 & $\mathcal{D}^{\prime}_3$                   & 84.72 & 80.24 & 82.41       & 48.87 & 55.91 & 52.14        & 54.07 & 41.99 & 47.21      & 81.41 & \textbf{81.48} & \textbf{81.42}       \\
                                 & $\mathcal{D}^{\prime}_4$                 & \textbf{85.71} & \textbf{81.16} & \textbf{83.37}       & \textbf{61.85} & \textbf{55.91} & \textbf{58.72}        & 55.00 & 42.05 & \textbf{47.61}      & -     & -     & -           \\
\hline
\multirow{4}{*}{RoBERTa-TopNeg}  & $\mathcal{D}$                             & 82.95                     & 78.56                     & 80.70                     & 52.77                     & 54.86                     & 53.80                     & 60.69                     & 45.21                     & 51.56                     & -                         & -                         & -                          \\
                                 & $\mathcal{D}^{\prime}_1$                         & 83.76 & 80.56 & 82.13       & 59.10 & 50.12 & 54.23        & 59.97 & 44.56 & 51.01      & -     & -     & -           \\  
                                 & $\mathcal{D}^{\prime}_2$                     & 85.50 & 77.28 & 81.18       & 55.10 & 54.31 & 54.69        & 58.26 & 50.11 & 53.83      & -     & -     & -           \\ 
                                 & $\mathcal{D}^{\prime}_3$                   & \textbf{86.74} & 81.40 & 83.98       & 51.15 & 58.57 & 54.56        & 60.20 & 48.01 & 53.28      & -     & -     & -           \\
                                 & $\mathcal{D}^{\prime}_4$                  & 86.34 & \textbf{83.17} & \textbf{84.72}       & \textbf{60.56} & \textbf{59.30} & \textbf{59.88}        & \textbf{58.07} & \textbf{51.38} & \textbf{54.45}      & -     & -     & -           \\
\hline
\end{tabular}
\end{adjustbox}
\vspace{-5pt}
\caption{Results of four base models trained on original and different versions of cleaned DS-NER datasets. $\mathcal{D}$ denotes the original distantly annotated data, and $\mathcal{D}^{\prime}_1$, $\mathcal{D}^{\prime}_2$, $\mathcal{D}^{\prime}_3$, and $\mathcal{D}^{\prime}_4$ to respectively represent the datasets cleaned using training dynamics from BERT, RoBERTa, BERT-TopNeg, and RoBERTa-TopNeg. A bold score indicates that the model trained on the corresponding data version achieves the best performance.}
\label{tab:cleaning_results}
\vspace{-7pt}
\end{table*}

\section{Results}
In this section, we report the results in two parts:
\S\ref{sec:noise_filtering}
for each dataset, we train different base models on the original distantly supervised data and on various cleaned versions of the data; and 
\S\ref{sec:comparison}
we compare the base models trained on our best-cleaned data with the baseline methods.

\subsection{Cleaning Results} \label{sec:noise_filtering}

In our cleaning experiments, we apply the four base models in our DynClean framework to clean each dataset.
Specifically, we use $\mathcal{D}$ to denote an original distant dataset, and $\mathcal{D}^{\prime}_1$, $\mathcal{D}^{\prime}_2$, $\mathcal{D}^{\prime}_3$, and $\mathcal{D}^{\prime}_4$ to respectively represent the datasets cleaned using training dynamics from BERT, RoBERTa, BERT-TopNeg, and RoBERTa-TopNeg. 
We then train the four base models on the original distant data and the different cleaned data.

Table \ref{tab:cleaning_results} gives the entire set of results.
From the results, we found that the model's performance consistently improved when it was trained with cleaned data compared to the performance with the original distant data.
Specifically, for CoNLL03, Wikigold, and WNUT16, models trained on $\mathcal{D}^{\prime}_4$ consistently achieve the best performance.
Therefore, we refer $\mathcal{D}^{\prime}_4$ as the best-cleaned data for these datasets.
For BC5CDR, it is $\mathcal{D}^{\prime}_3$.
Particularly, RoBERTa trained by best-cleaned Wikigold ($\mathcal{D}^{\prime}_4$) shows an 8.67\% improvement in F1 score when compared with original distant data.
We also have the following observations: \ding{182} For each dataset, the model used in DynClean to obtain the best-cleaned data also has the best performance in the original distant data.
RoBERTa-TopNeg, which obtains the best-cleaned data for CoNLL03, Wikigold, and WNUT16, also performs the best on their original distant data.
For BC5CDR, the model is BERT-TopNeg.
We believe this is because, when the model used in DynClean has better performance on the original distant data, the obtained training dynamics  are more capable to reflect the ``true'' data sample characteristics for cleaning;
\ding{183} When models are trained on the best-cleaned data, it is not necessary to employ TopNeg.
For the best-cleaned CoNLL03 ($\mathcal{D}^{\prime}_4$), WNUT16 ($\mathcal{D}^{\prime}_4$), and BC5CDR ($\mathcal{D}^{\prime}_3$), models trained on them without TopNeg achieve better performances.
BERT is the outlier, when trained  with TopNeg on the WikiGold ($\mathcal{D}^{\prime}_4$) it exhibits a \textit{slight} improvement.
This indicates that our approach removes a large enough amount of false negatives, which obviates the necessity for employing additional negative sampling techniques.
Our additional experiments in Appendix \ref{apx:HA_exp} also show that TopNeg is not necessary when training models on the \textit{human-annotated} data.

\subsection{Baselines Comparison} \label{sec:comparison}

\begin{table*}[h!]
\small
\centering
\begin{adjustbox}{width=1\textwidth}
\begin{tabular}{lcccccccccccc} 
\hline
\multirow{2}{*}{Methods} & \multicolumn{3}{c}{CoNLL03}                      & \multicolumn{3}{c}{WikiGold}                     & \multicolumn{3}{c}{WNUT16}                       & \multicolumn{3}{c}{BC5CDR}                        \\ 
\cline{2-13}
                         & P              & R              & F1             & P              & R              & F1             & P              & R              & F1             & P              & R              & F1              \\ 
\hline
\textbf{Fully Supervised} \\
RoBERTa           & 90.05              & 92.48              & 91.25          & 85.33             &  87.56             & 86.43         & 51.76              &  52.63             & 52.19         &  88.41             &  87.28             & 89.56          \\ 
BiLSTM-CRF           &  89.14              & 91.10              & 90.11         & 55.40         & 54.30            &  54.90         & 60.01             &  46.16             & 52.18         &  89.09             &  62.22             & 73.27         \\ 
\hline
\textbf{LLMs Evaluation} \\
Llama-3.1-70B           & 71.08              & 83.37           & 76.74          &  57.67            &  48.15            & 52.48         &  50.00             &  45.46           & 47.62        & 67.23            & 55.87           &  61.03         \\ 
ChatGPT           & 74.29             & 84.23             & 78.95          &  63.51             &  48.61             & 55.07         &  54.55             &  50.00           & 52.17         & 80.61             & 68.91              & 74.30           \\ 
\hline
\textbf{Distantly Supervised} \\
KB-Matching            & 81.13          & 63.75          & 71.40          & 47.90          & 47.63          & 47.76          & 40.34          & 32.22          & 35.83          & 86.39          & 51.24          & 64.32           \\
AutoNER$^{\ddagger}$             & 60.40          & 75.21          & 67.00          & 52.35          & 43.54          & 47.54          & 18.69          & 43.26          & 26.10          & 77.52          & 82.63          & 79.99           \\
bnPU $^{\dagger}$                  & 82.97          & 74.38          & 78.44          & -              & -              & -              & -              & -              & -              & 77.06          & 48.12          & 59.24           \\
BOND$_{\textrm{RoBERTa}}$       & 83.76          & 68.90          & 75.61          & 49.17          & 54.50          & 51.55          & 53.11          & 41.52          & 46.61          & -              & -              & -               \\
CReDEL                 & -              & -              & -              & -              & -              & -              & -              & -              & -              & 71.70          & \textbf{86.80} & 78.60           \\
Span-NS  $^{\ddagger}$                & 80.41          & 71.35          & 75.61          & 51.05          & 48.27          & 49.62          & 53.51          & 39.76          & 45.62          & \textbf{86.90} & 73.49          & 79.64           \\
Span-NS-V $^{\ddagger}$              & 80.19          & 72.91          & 76.38          & 50.91          & 48.43          & 49.64          & 47.78          & 44.37          & 46.01          & 86.67          & 73.52          & 79.56           \\
BERT-TopNeg$^{\ddagger}$  & 82.72          & 77.71          & 80.08          & 55.47          & 48.57          & 50.65          & 55.28          & 40.35          & 46.55          & 82.09          & 78.90          & 80.39           \\
RoBERTa-TopNeg$^{\ddagger}$        & 81.07          & 80.23          & 80.55          & 52.30          & 53.55          & 52.86          & \textbf{60.55} & 45.33          & 51.78          & -              & -              & -               \\ 
Conf-MPU$_{\textrm{LBiLSTM}}$           & 77.39          & \textbf{82.84}          & 80.02          & -              & -              & -              & -              & -              & -              & 76.63          & 83.82          & 80.07           \\
Conf-MPU$_{\textrm{BERT}}$         &    78.58      &  79.75          & 79.16          & -              & -              & -              & -              & -              & -              &   69.79         &    86.42      & 77.22           \\
MProto$_{\textrm{BERT}}$         &    79.80      &  79.37          & 79.58          & -              & -              & -              & -              & -              & -              &   77.53         &    85.84      & 81.47           \\
\hline
\textbf{Ours}$_{\textrm{BERT}}$        &   87.29       &  80.40      &    83.70        &   60.75  &     56.71      &     58.65      &      58.75     &    40.25       &    47.74      &     80.02      &     83.06      &   \textbf{81.50}  \\
\textbf{Ours}$_{\textrm{RoBERTa}}$      & \textbf{88.41} & 82.77 & \textbf{85.50} &  \textbf{60.94}     & \textbf{59.31} & \textbf{60.01} & 58.22         &  \textbf{51.35} & \textbf{54.52} & -              & -              & -               \\
\hline
\end{tabular}
\end{adjustbox}
\caption{Comparisons with baselines on four datasets. 
$^{\dagger}$ marks the results retrieved from \citet{zhou-etal-2022-distantly} and $^{\ddagger}$ marks the results from \citet{xu-etal-2023-sampling}. The best results are in \textbf{bold}.
}
\label{tab:main_results}
\vspace{-10pt}
\end{table*}

The comparison against the baselines is summarized in Table \ref{tab:main_results}. 
As our method, we present the results of BERT and RoBERTa trained on the best-cleaned version of each dataset, i.e., the CoNLL03 ($\mathcal{D}^{\prime}_4$), Wikigold ($\mathcal{D}^{\prime}_4$), WNUT16 ($\mathcal{D}^{\prime}_4$), and BC5CDR ($\mathcal{D}^{\prime}_3$).
We first observe that LLMs still face certain challenges in the NER task, with a significant performance gap compared to fully supervised RoBERTa. 
This result is similar to previous study \citep{qin2023chatgpt}, indicating that LLMs still face significant challenges in NER task.
Our approach also outperforms all LLMs across four datasets.
We further found that LLMs outperform KB-Matching, indicating the feasibility of using LLMs for distant annotation. We have provided a related discussion in Appendix \ref{apx:LLMs_distant}.
Despite using \textit{less} data for model training, we outperform all previous approaches with relatively balanced precision and recall scores.
When comparing with Conf-MPU$_{\textrm{BERT}}$, BERT trained on the cleaned CoNLL03 and BC5CDR achieves an F1 score improvement of 4.54\% and 4.28\%, respectively.
Our method also outperforms Conf-MPU$_{\textrm{LBiLSTM}}$, which is enhanced with lexicon feature engineering, with the F1 score improvement of 3.68\% and 1.43\%, respectively.
Compared to MProto$_{\textrm{BERT}}$, we have 4.12\% F1 improvement on the CoNLL03 dataset.
As for the CReDEL which also focuses on enhancing the data quality of distantly supervised datasets, our approach outperforms it by 2.9\% F1 score on the BC5CDR dataset.
Our approach also outperforms multiple baselines that utilize negative sampling techniques.
The main reason our method achieves significant improvement is that it is able to remove a large number of false negatives and  positives from the distantly annotated data.
We further test the effectiveness of using our cleaned datasets with the self-training method, and the results can be found in Appendix \ref{apx:self_train}.

\section{Analysis and Discussion}
In this section, we provide the ablation study and case studies to better understand the effectiveness of our proposed method.
In addition, we provide experimental results of utilizing another frequently used training dynamics metric in our DynClean.

\subsection{Ablation Study}
We conduct the ablation study on two aspects: \ding{182} cleaning either negative samples or positive samples, but not both; \ding{183} varying the percentile in threshold samples for computing positive and negative thresholds.
All results are reported from experiments on cleaned datasets with training dynamics from RoBERRTa-TopNeg for cleaning. 
Due to space limitations, we provide the analysis of CoNLL03 in this section. The ablation studies for the remaining datasets are included in Appendix \ref{more_ablation}.

\noindent\textbf{Cleaning positive/negative only.}\quad
Table \ref{tab:ab_fnfp} shows the performance when only cleaning either negative samples or positive samples.
We observe that if we single out the negative samples in cleaning, we improve the recall by a large margin, but with a notable loss in precision.
Conversely, focusing on cleaning positive samples yields high precision but very low recall.
This demonstrates the necessity of simultaneously removing both types of noisy annotation to create better distant labels.

\begin{table}
\footnotesize
\centering
\begin{tabular}{lccc} 
\hline
  Training Data    & P     & R     & F1     \\ 
\hline
$\mathcal{D}^{\prime}$ & 88.41 & 82.77 & 85.50  \\ 
\hline
$\mathcal{D}^{\prime}_{\textrm{neg}}$    &   75.04 &  83.76  & 79.16 \\
$\mathcal{D}^{\prime}_{\textrm{pos}}$    & 93.03 & 64.94 & 76.49  \\
$\mathcal{D}$   & 90.16 & 65.88 & 76.13  \\
\hline
\end{tabular}
\vspace{-3pt}
\caption{
Results of only cleaning either negative samples or positive samples.
$\mathcal{D}^{\prime}_{\textrm{pos}}$ and $\mathcal{D}^{\prime}_{\textrm{neg}}$ denote denote only clean positive and negative, respectively.
$\mathcal{D}^{\prime}$ denotes data cleaned both and $\mathcal{D}$ is the original distant data.
}
\label{tab:ab_fnfp}
\vspace{-15pt}
\end{table}

\begin{figure*}[t]
    \centering
    \includegraphics[width=1\textwidth]{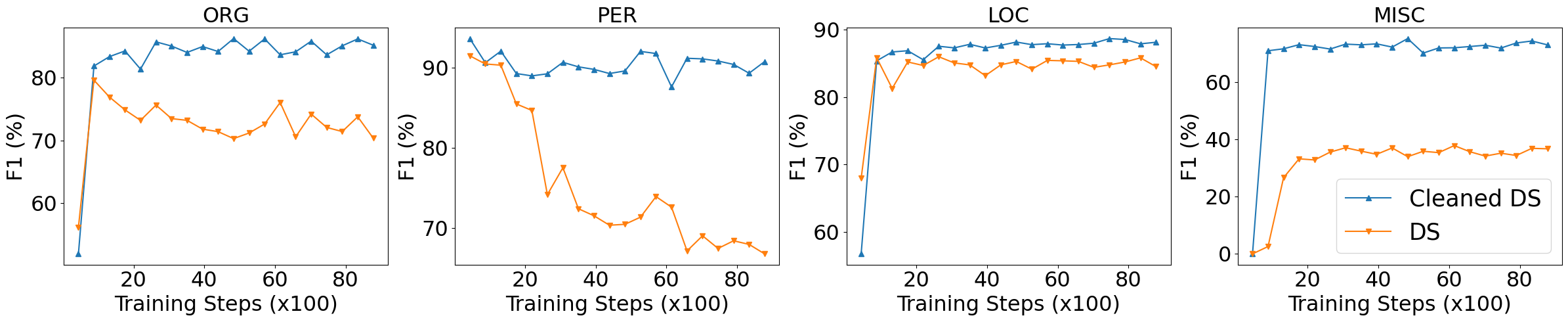}
    \vspace{-15pt}
    \caption{The performance curve for each class of CoNLL03 when training with the original DS and cleaned DS $\mathcal{D}^{\prime}_4$ and testing on the dev set. ``ORG'', ``PER'', ``LOC'', and ``MISC'' represent the entity types of organization, person, location, and miscellaneous, respectively. 
    }
    \label{fig:curve}
    \vspace{-15pt}
\end{figure*}

\begin{figure}[tb]
   \begin{minipage}{0.24\textwidth}
     \centering
     \includegraphics[width=\linewidth]{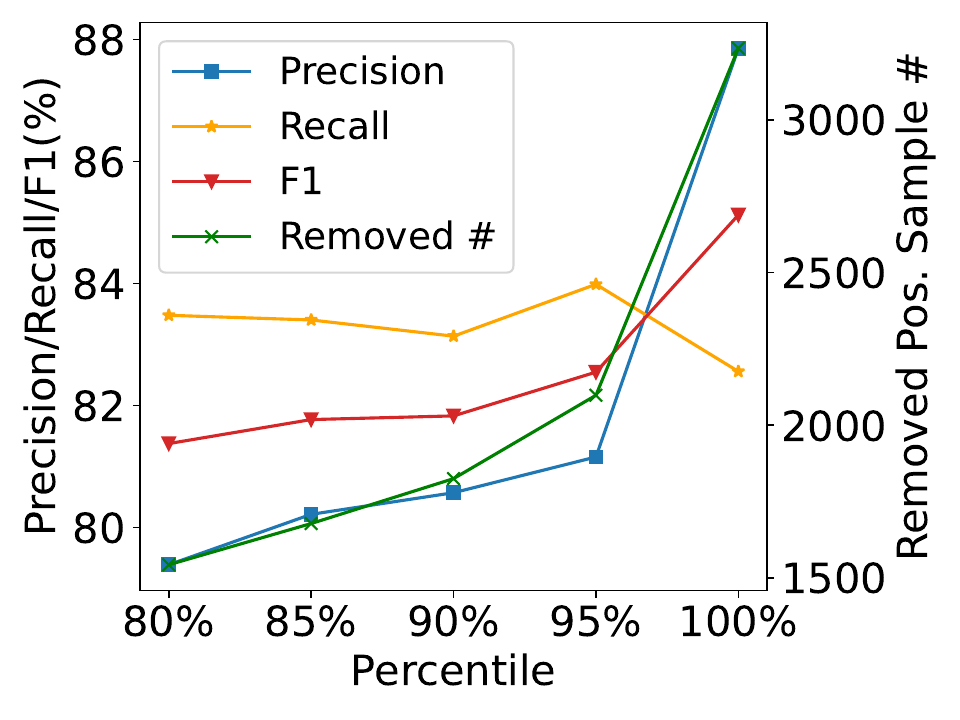}
   \end{minipage}\hfill
   \begin{minipage}{0.24\textwidth}
     \centering
     \includegraphics[width=\linewidth]{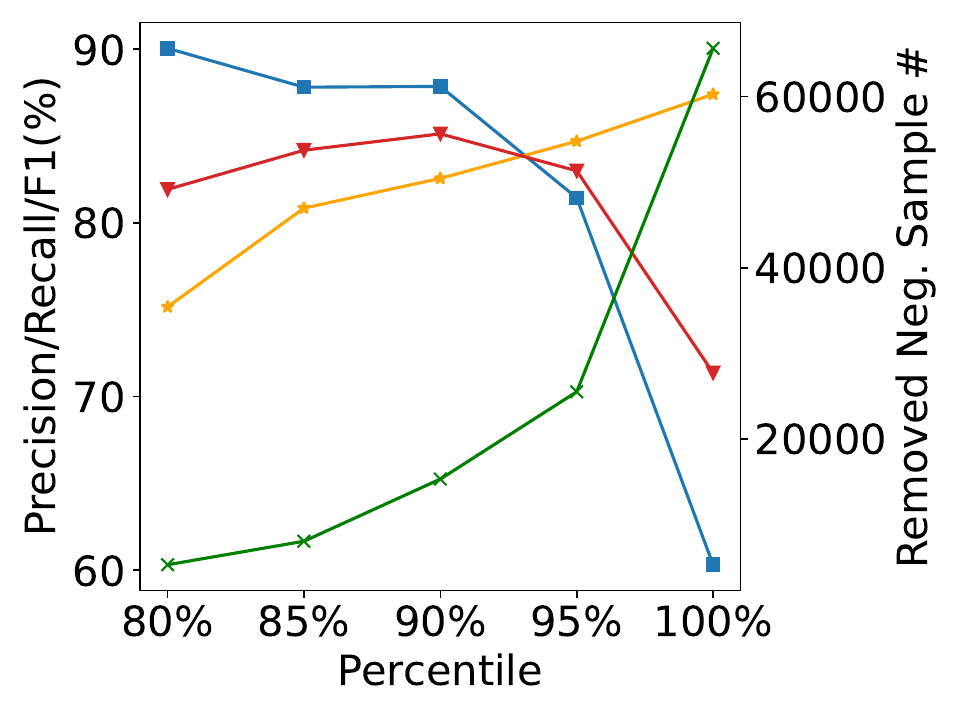}
   \end{minipage}
\vspace{-10pt}
\caption{Ablation of varying percentile in threshold samples to compute the thresholds. 
}
\label{fig:percentile_ablation}
\end{figure}

\noindent\textbf{Varying threshold sample percentile.}\quad 
The optimal threshold should eliminate as many false samples as possible while minimizing the removal of true samples.
Figure \ref{fig:percentile_ablation} presents the performance and the number of discarded samples associated with various percentiles. 
The left plot illustrates the results when the percentile for computing the negative sample threshold is fixed at 90\%, and the percentile for positive threshold samples is varied. 
The right plot shows the trends when the percentile of positive threshold samples is fixed at 100\%, and the negative sample percentile is varied.
For positive samples, both the precision and F1 scores show a consistent improvement.
Regarding negative samples, though the recall continues to increase, the F1 score decreases when the percentile exceeds 90\%. 
We also note that the number of discarded negative samples has a more pronounced increase as the percentile increases, which gives a significantly higher removal.
The above empirical observation suggests that 
when the percentile exceeds 90\% denoising becomes too aggressive in eliminating the false negatives which comes at the expense of discarding a large number of true negatives, leading to a decline in the overall performance.
The true samples that exhibit training dynamics similar to false samples, are deemed \textit{hard} samples \citep{talukdar-etal-2021-training}.

\subsection{Effective of Noise Cleaning}
We conducted a further study to evaluate the noise cleaning effectiveness of our proposed DynClean.
We use the original DS version and cleaned DS version of $\mathcal{D}^{\prime}_4$ of CoNLL03 to train the RoBERTa and plot the class-wise performance curve on development set, as shown in Figure \ref{fig:curve}.
We found that the model trained on our cleaned DS data consistently outperforms across classes than the model trained on original DS data, though we used less (but of higher-quality) training data. 
For the ORG and PER, we observed a gradual decrease in the model's performance when trained on the original DS dataset. This decline is attributed to the model overfitting on the extensive amount of noisy data. In contrast, training on the cleaned dataset allows the model to converge better.
Notably, we find a significant improvement in the MISC.
We provide in-depth for it in Appendix \ref{apx:cleaning_effect}.
Specifically, since the MISC has the lowest proportion and the DS annotation results a large number of false negative annotation in MISC (i.e., missing MISC spans).
Our cleaning method removed a substantial number of false negative spans of MISC, resulting in a significant performance improvement.
We provide additional noise cleaning analysis on the WikiGold dataset in Appendix \ref{sec:wikigold_cleaning}, as it represents a more complex DS-NER scenario.

\subsection{Qualitative Examples}\label{apx:qualitative}

\begin{table*}[tb]
\small
\centering
\begin{adjustbox}{width=1\textwidth}
\begin{tabular}{l}
\hline
\textbf{DS instance}:~\textcolor{blue}{[Union]$_{ORG}$} officials from the \textcolor{blue}{[Public Service Association]$_{ORG}$} ( PSA ) were unavailable for comment.  \\
\textbf{HA instance}:~Union officials from the \textcolor{blue}{[Public Service Association]$_{ORG}$} ( \textcolor{blue}{[PSA]$_{ORG}$} ) were unavailable for comment.  \\
\textbf{Ours}:~ \textcolor{red}{[\st{Union}]$_{FP}$} officials from the \textcolor{blue}{[Public Service Association]$_{ORG}$} ( \textcolor{red}{[\st{PSA}]$_{FN}$} ) were unavailable for comment. \\
\hline
\textbf{DS instance}:~Orioles's bench coach \textcolor{green}{[Andy]$_{PER}$} Etchebarren will manage the club in \textcolor{green}{[Johnson]$_{PER}$}'s absence. \\
\textbf{HA instance}:~\textcolor{blue}{[Orioles]$_{ORG}$}'s bench coach \textcolor{green}{[Andy Etchebarren]$_{PER}$} will manage the club in \textcolor{green}{[Johnson]$_{PER}$}'s absence. \\
\textbf{Ours}: ~\textcolor{red}{[\st{Orioles}]$_{FN}$}'s bench coach \textcolor{red}{[[\st{Andy}]$_{FP}$ \st{Etchebarren}]$_{FN}$} will manage the club in \textcolor{green}{[Johnson]$_{PER}$}'s absence. \\
\hline
\end{tabular}
\end{adjustbox}
\caption{Case study for the CoNLL03 dataset. ``DS'' denotes the distantly annotated instances and ``HA'' denotes the Human-Annotated instances. The {\color[HTML]{EA3323}FN} and {\color[HTML]{EA3323}FP} are the identified false negative and false positive annotations in DS data by our method, which are removed during training.}
\label{tab:case_study}
\end{table*}

To give an intuition of the benefits of our proposed false annotation cleaning approach for DS-NER, we present qualitative examples in Table \ref{tab:case_study}. It has two examples.
One observes that compared to the HA (Human-Annotated) instance, the DS (Distant Supervision) instance omits the entity spans ``PSA'', ``Orioles'', and ``Andy Etchebarren'' (false negatives), incorrectly labels ``Union'' and partially labels ``Andy'' as entity spans (false positive).
Our method successfully identifies both false negative and false positive annotations in the DS data.
Avoiding such false samples for training can alleviate the models' overfitting noise.
This also indicates that our method can assist the distant annotation process and improve the data quality.

\subsection{Other Training Dynamics Metric} \label{sec:confidence}
We evaluate another frequently used training dynamic metrics defined as mean (Confidence) and standard deviation (Variability) of the probabilities estimated by the model for a given label across epochs \citep{swayamdipta-etal-2020-dataset}.
We provide detailed definitions of both in Appendix \ref{apx:conf_define}.
The data are then categorized as \textbf{easy-to-learn}, \textbf{ambiguous}, and \textbf{hard-to-learn} according to the two metrics. 
It is hypothesized that the \textbf{hard-to-learn} region usually contains mislabeled samples \citep{swayamdipta-etal-2020-dataset}.
We apply the same threshold estimation strategy as proposed in Section \ref{sec:denoising_method}
to estimate the Confidence thresholds for filtering out noisy samples.
Specifically, the Eq. \ref{eq:AUM} in Algorithm \ref{alg:noise_filter} is replaced by Eq. \ref{eq:confidence}.
Table \ref{tab:other_metric} presents the experimental results for the two metrics by using RoBERTa-TopNeg as the base model for both cleaning and evaluating.
We observe that the two metrics exhibit very similar performance.
This is because samples with low AUM values also have relatively low confidence scores, indicating they are \textbf{hard-to-learn} samples.
Figure \ref{fig:datamap_conll03} in Appendix \ref{apx:datamap} demonstrates a significant overlap between samples with low AUM values and those with low confidence values.

\begin{table}[t]
\footnotesize
\begin{adjustbox}{width=0.48\textwidth}
\centering
\begin{tabular}{ccccc} 
\hline
           & CoNLL03 & WikiGold & WNUT16 & BC5CDR  \\ 
\hline
AUM        & 85.50   &  60.01    & 54.45  & 81.50   \\
Conf. & 83.35   & 58.55    & 52.62  & 80.16   \\
\hline
\end{tabular}
\end{adjustbox}
\vspace{-5pt}
\caption{Comparative F1-score results of AUM-based and Confidence-based (``Conf.'') training dynamics.
}
\label{tab:other_metric}
\vspace{-15pt}
\end{table}

\section{Conclusion}
In this paper, we propose DynClean, a training dynamics-based cleaning approach for distantly supervised NER tasks.
Unlike most existing methods that focus on learning from noisy labels, DynClean aims to improve the quality of data generated by distant supervision annotation.
DynClean leverages the model behavior on each sample during the training to characterize samples, thereby locating both false positive and false negative annotations.
Extensive experiment results show that models trained on our cleaned datasets achieve improvement ranging from 3.19\% to 8.95\% in F1-score; it also outperforms SOTA  DS-NER works by significant margins, up to 4.53\% F1-score, despite using fewer samples in training.

\section*{Limitations}
Our method employs a span-based NER model, which has lower inference efficiency compared to token-based NER models.
Although our proposed DynClean method achieves better performance than more sophisticated approaches, DynClean has high performance requirement on the model used for calculating accurate training dynamics, which may increase the computational cost.
Considering the effectiveness of performance improvement, we believe the additional computational costs are acceptable.
Developing strategies to reduce the performance requirements for cleaning will increase the applicability of our method.
Additionally, 
while our method successfully identifies numerous false samples, it may also inadvertently discard correctly labeled but hard samples.
A future research direction involves refining our approach to better distinguish between false and hard samples.
Furthermore, employing a noise-robust loss function on identified ``noisy'' samples may enhance model performance than simply removing them.

\section*{Acknowledgements}
This work was supported by the National Science Foundation awards III-2107213, III-2107518, and ITE-2333789.
We also thank our reviewers for their insightful feedback and comments.


\bibliography{anthology,custom}
\clearpage
\appendix

\section{Datasets statistics} \label{apx:dataset_stat}

Table \ref{tab:dataset_stat} presents the statistics of the used four datasets.
``\# Sent.'' denotes the number of sentences and ``\# Entity'' indicates the number of entities (positive span samples) in the datasets.
All the training sets are annotated by distant supervision, and the development and test sets are human annotated.
The entity types number of CoNLL03, WikiGold, WNUT16, and BC5CDR are 4, 4, 10, and 2.
CoNLL03 and WikiGold are the general domain NER.
WNUT16 is open domain NER dataset and
BC5CDR is biomedical domain NER dataset.

\begin{table}[ht]
\small
\centering
\begin{adjustbox}{width=0.48\textwidth}
\begin{tabular}{cccccc} 
\hline
Datasets               &           & CoNLL03 & WikiGold & WNUT16 & BC5CDR  \\ 
\hline
\multirow{2}{*}{Train} & \# Sent.  & 14,041  & 1,142    & 2,393  & 4,560   \\
                       & \# Entity & 17,781  & 2,282    & 994    & 6,452   \\ 
\hline
\multirow{2}{*}{Dev}   & \# Sent.  & 3,250   & 280      & 1,000  & 4,579   \\
                       & \# Entity & 5,942   & 648      & 661    & 9,591   \\ 
\hline
\multirow{2}{*}{Test}  & \# Sent.  & 3,453   & 274      & 3,849  & 4,797   \\
                       & \# Entity & 5,648   & 607      & 3,473  & 9,809   \\
\hline
\end{tabular}
\end{adjustbox}
\caption{Statistics of four DS-NER datasets.}
\label{tab:dataset_stat}
\end{table}

\begin{figure}[h!]
\centering
\includegraphics[width=\linewidth]{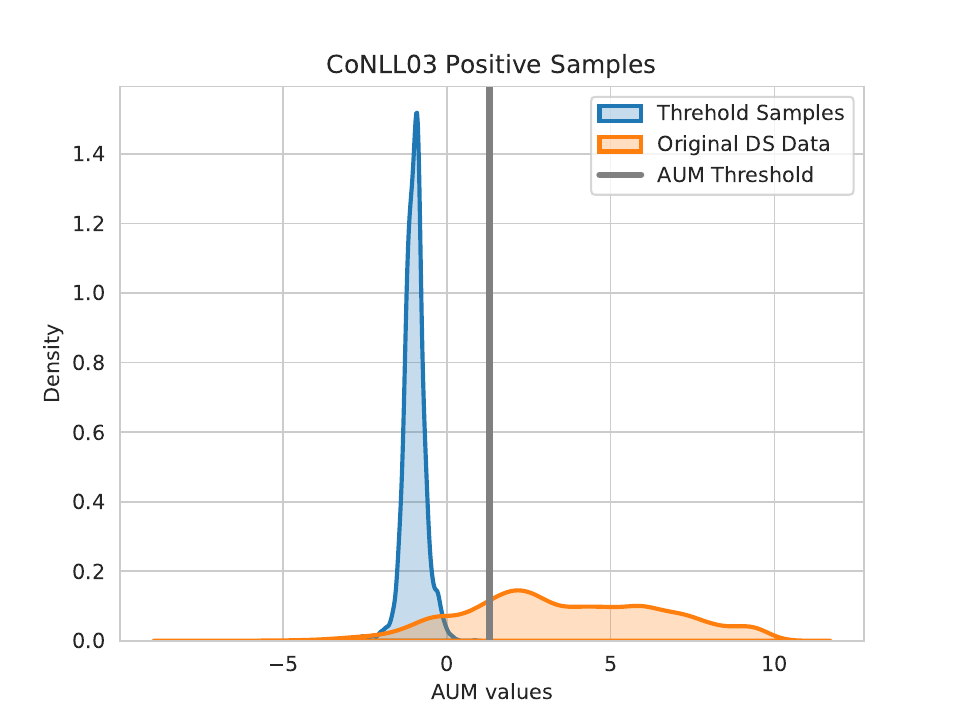}
\caption{AUM distributions of positive samples and positive threshold samples.}
\label{fig:aum_pos_thres}
\end{figure}

\begin{figure}[h!]
\centering
\includegraphics[width=\linewidth]{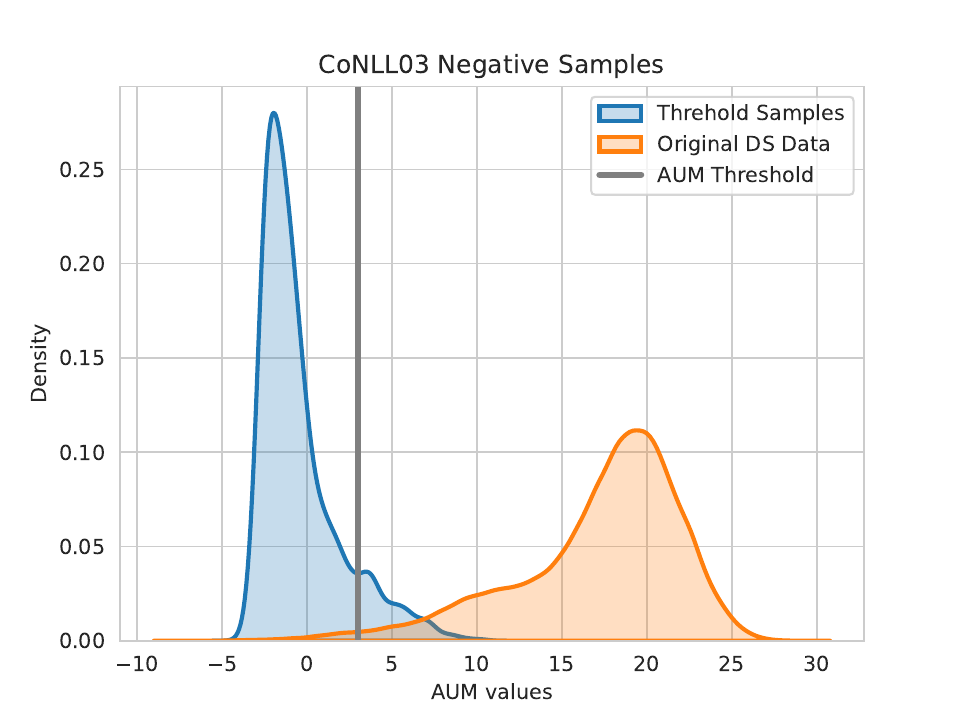}
\caption{AUM distributions of negative samples and negative threshold samples.}
\label{fig:aum_neg_thres}
\end{figure}

\begin{table}[ht]
\small
\begin{adjustbox}{width=0.48\textwidth}
\centering
\begin{tabular}{ccccc} 
\hline
\multirow{2}{*}{Dataset} & \multicolumn{2}{c}{\begin{tabular}[c]{@{}c@{}}AUM $k$~\%ile in~\\threshold samples\end{tabular}} & \multicolumn{2}{c}{\begin{tabular}[c]{@{}c@{}}AUM \%ile in \\DS data\end{tabular}}  \\ 
\cline{2-5}
                         & Positive & Negative                                                                  & Positive & Negative                                                                 \\ 
\hline
CoNLL03                  & 100\%    & 90\%                                                                      & 26.7\%   & 1.5\%                                                                    \\
WikiGold                 & 100\%    & 90\%                                                                      & 28.3\%   & 0.5\%                                                                    \\
WNUT16                   & 100\%    & 90\%                                                                      & 9.8\%    & 0.1\%                                                                    \\
BC5CDR                   & 90\%     & 90\%                                                                      & 1.3\%    & 0.9\%                                                                    \\
\hline
\end{tabular}
\end{adjustbox}
\caption{The percentile of the threshold values obtained through our proposed method in the threshold samples, and the corresponding percentile of the threshold values in the DS data.}
\label{tab:thresholdStat}
\end{table}

\section{Effectiveness of threshold estimation strategy} \label{apx:thre_estimate_analysis}

Estimating the thresholds using threshold samples proves more consistent than direct tuning on the original DS data.
Figure \ref{fig:aum_pos_thres} and \ref{fig:aum_neg_thres} illustrates the distribution of AUM values for both threshold samples and original DS samples in the CoNLL03 dataset.
The gray lines in the figure represent the AUM threshold values corresponding to the  $k_{\textrm{pos}}$ and $k_{\textrm{neg}}$ percentiles in threshold samples.
Subsequently, samples with values lower than these threshold values (i.e., below the gray lines) in the original DS-NER dataset are eliminated during model training.
In this case, the tuned percentiles are 100\% for positive and 90\% for negative in threshold samples.
However, if tuning is performed directly on the original DS-NER dataset, the corresponding percentiles are 26.7\% for positive and 1.5\% for negative.
Compared to those in threshold samples, which mimic the training dynamics of mislabeled samples, these percentiles are more challenging to obtain.
Table \ref{tab:thresholdStat} shows the corresponding percentiles of threshold values in threshold samples and original DS samples for all tested datasets.
It is observed that the percentile selection in our method exhibits greater consistency.
In contrast, the corresponding percentiles in the original DS-NER datasets are very dataset-dependent.
This variation arises due to the differing noise distributions in DS-NER datasets, which complicates the direct tuning of percentile choices in the original datasets.

\section{Additional Experimental Settings} \label{apx:addional_settings}
\subsection{Additional Experimental Settings of DynClean}
For a fair comparison, we use the same versions of the encoders in the above baselines.
\textit{bert-base-cased}\footnote{https://huggingface.co/bert-base-cased} as BERT, \textit{roberta-base}\footnote{https://huggingface.co/roberta-base} as RoBERTa, and \textit{biobert-base-cased-v1.1}\footnote{https://huggingface.co/dmis-lab/biobert-v1.1} as BioBERT.
For all tested span-based NER models, we use the same combnition of hyperparameters for all datasets: the learning rate is set as 1e-5; the training batch size is 16, and the maximum span length $L$ is set as 8; 2 layers of feed-forward neural networks (FFNN) is employed as the classifier and the hidden size is set as 150, and the dropout rate is set as 0.2; the learnable width embedding size is 150.
When using the TopNeg, we set the top-$N_r$ as 5\% as suggested by \citep{xu-etal-2023-sampling}.
For CoNLL03, BC5CDR, the training epoch $E$ in \ref{alg:noise_filter} is 5. For WikiGold and WNUT16, the training epoch $E$ is 10.
Except for AutoNER, all other DS-NER baselines use the dev set for hyperparameter tuning. We follow this setting to tune the $k_{\textrm{neg}}$ and $k_{\textrm{pos}}$ for each dataset.
Experiments were conducted using a single NVIDIA RTX 8000 GPU card with PyTorch 2.10.0, and the reported results represent the average of five runs, each with a different random seed.
The average running time on the CoNLL03 dataset is 78 seconds/epoch.

\subsection{Experimental Settings of LLMs}

\begin{figure}[t!]
\centering
\includegraphics[width=\linewidth]{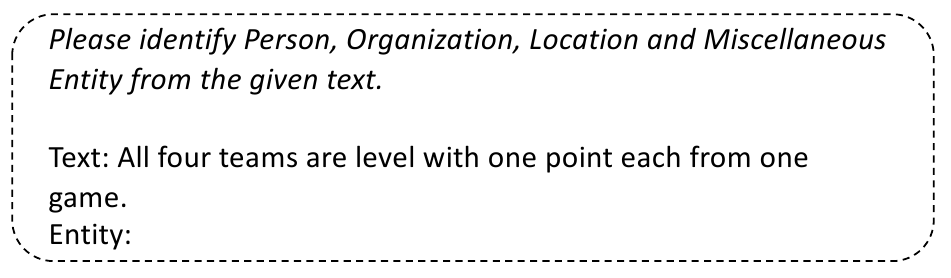}
\vspace{-10pt}
\caption{The prompts and input formats for our experiments. The shown example is based on the entity types of CoNLL03 dataset.}
\label{fig:prompt}
\end{figure}

We implement the ChatGPT (\textit{gpt-3.5-turbo}) and LLama-3.1-70B with zero-shot setting as in the distantly supervised NER task without any access to human annotated training label. 
The Llama-3.1-70B is ran on two A100 80G GPUs and uses the exactly same prompts of ChatGPT.
For each dataset, we randomly sampling 200 samples from the text set each time. Our experiments were repeated five times and the results were averaged. We follow the prompt setting used in \citep{qin2023chatgpt} as illustrated in Figure \ref{fig:prompt}.

For the Distantly-Annotation from ChatGPT, we tested BC5CDR and CoNLL03. The tottal cost of API usage is 20.58 dollars.

\section{Impact of TopNeg on Human Annotated Data} \label{apx:HA_exp}

\begin{table}[t]
\small
\begin{adjustbox}{width=0.48\textwidth}
\centering
\begin{tabular}{clccc} 
\hline
Training Data       & Model          & P     & R     & F1     \\ 
\hline
\multirow{4}{*}{HA} & BERT           & 89.71 & 90.32 & 90.01  \\
                    & BERT-TopNeg    & 84.25 & 92.23 & 88.06  \\
                    & RoBERTa        & 90.05 & 92.48 & 91.25  \\
                    & RoBERTa-TopNeg & 87.81 & 92.74 & 90.21  \\
\hline
\multirow{4}{*}{DS} & BERT           & 89.75 & 63.08 & 72.09  \\
                    & BERT-TopNeg    & 79.42 & 79.77 & 79.58  \\
                    & RoBERTa        & 90.16 & 65.88 & 76.13  \\
                    & RoBERTa-TopNeg & 82.95 & 78.56 & 80.7  \\
\hline
\multirow{4}{*}{Ours} & BERT         & 87.29 & 80.40 & 83.70  \\
                    & BERT-TopNeg    & 85.71 & 81.16 & 83.37  \\
                    & RoBERTa        & 87.64 & 82.67 & 85.08  \\
                    & RoBERTa-TopNeg & 86.34 & 83.17 & 84.72  \\
\hline
\end{tabular}
\end{adjustbox}
\caption{Comparisons on Human-annotated (HA), Distantly Supervised (DS), and our cleaned data of CoNLL03 for training.}
\label{tab:topneg_ha}
\end{table}

In Table \ref{tab:topneg_ha}, we show the results of training the span-based NER models with and without TopNeg on the Human-annotated (HA), Distantly Supervised (DS), and our best-cleaned CoNLL03 ($\mathcal{D}^{\prime}_4$) datasets.
We can observe that better performance is obtained when not using TopNeg on HA data.
When training models on our denoised DS data, without TopNeg perform better as well.
But TopNeg improves the models on the DS data.
This indicates that our distant label cleaning method removes a large enough amount of false negatives.

\section{LLMs as Distant Annotator} \label{apx:LLMs_distant}

We used the predictions of the open-source Llama-3.1-70B on CoNLL03 as distant labels. 
We follow the experimental settings in Section \ref{sec:comparison} to train RoBERTa and our DynClean using this data. 
The experimental results are shown in Table \ref{tab:llm_distant}. 
We found that, because Llama-3.1-70B outperformed KB-Matching, our model achieved further improvements. This demonstrates the feasibility of combining our method with LLM predictions as distant labels to enhance the DS-NER performance.

\begin{table}[t]
\small
\centering
\begin{adjustbox}{width=0.4\textwidth}
\begin{tabular}{lccc}
\hline
              & P     & R     & F1    \\ \hline
Llama-3.1-70B & 71.08 & 83.37 & 76.74 \\ \hline
RoBERTa       & 86.84 & 76.04 & 81.07 \\
\textbf{Ours}$_{\textrm{RoBERTa}}$  & 92.41 & 85.05 & 88.58 \\ \hline
\end{tabular}
\end{adjustbox}
\caption{Results of DynClean and RoBERTa trained on the distant labels from Llama-3.1-70B of CoNLL03 dataset.}
\label{tab:llm_distant}
\end{table}

\section{Applying Self-Training on Cleaned Data} \label{apx:self_train}

In this section, we study the integration of our proposed data cleaning approach with self-training method to enhance the overall performance of NER.
We apply the self-training framework the same as \cite{KDDDSNER} on the best-cleaned version of each dataset. 

The results are shown in Table \ref{tab:self_train}.
As demonstrated in the Table, applying self-training to our label-cleaned dataset led to further performance improvements across all datasets. 
Our approach achieved performance parity with DesERT on both CoNLL03 and WikiGold. 
Moreover, we observed even more improvements on the remaining two datasets compared with DesERT. Notably, DesERT employs a more complex framework including two pre-trained language models (RoBERTa-base and DistilRoBERTa) and several sophisticated components (double-head pathway, dual co-guessing mechanism, worst-case cross-entropy loss and joint prediction) \cite{wang2024debiased}. 
These structures also increase the difficulty of training the model. In contrast, our self-training uses just a streamlined self-training approach with a single RoBERTa-base model and standard cross-entropy loss as described in Section \ref{Method:NER_model}. When reproducing their results, we found that we could not fully achieve their reported performance. Comparing with DesERT*, ours achieved better performance on three datasets (Wikigold, WNUT16, and BC5CDR).

\begin{table}[t]
\small
\begin{adjustbox}{width=0.48\textwidth}
\centering
\begin{tabular}{lcccc}
\hline
\textbf{Method} & \textbf{CoNLL03} & \textbf{Wikigold} & \textbf{WNUT16} & \textbf{BC5CDR} \\
\hline
BOND          & 81.48 & 60.07 & 48.01 & 75.60 \\
DesERT        & 86.95 & 65.99 & 52.26 & -     \\
DesERT*       & 86.72 & 64.49 & 51.24 & 80.21 \\
Ours          & 85.50 & 60.01 & 54.52 & 81.50 \\
Ours-ST       & 86.68 & 65.59 & 55.03 & 81.79 \\
\hline
\end{tabular}
\end{adjustbox}
\caption{Comparisons of models trained on our cleaned dataset (``Ours'') and applying self-training for further improvement (``Ours-ST'') with other advanced self-training based methods. BOND refers \cite{KDDDSNER}, DesERT refers \cite{wang2024debiased} and DesERT* refers the results we reproduced.}
\label{tab:self_train}
\end{table}



\section{Ablation Study on Other Datasets} \label{more_ablation}

\noindent\textbf{Cleaning positive/negative only.}\quad
Table \ref{tab:more_ablation} shows the performance when only cleaning either negative samples or positive samples for each remaining dataset.
The results demonstrate the necessity of simultaneously removing both types of noisy annotation to create better distant labels for each dataset.

\begin{table*}[htb] 
\small
\centering
\begin{adjustbox}{width=0.8\textwidth}
\begin{tabular}{lccccccccc}
\hline
              & \multicolumn{3}{c}{WikiGold} & \multicolumn{3}{c}{WNUT16}  & \multicolumn{3}{c}{BC5CDR}  \\ \cline{2-10} 
Training Data & P        & R       & F1      & P       & R       & F1      & P       & R       & F1      \\ \hline
$\mathcal{D}^{\prime}$             & 60.94    & 59.31   & 60.09   & 58.22   & 51.35   & 54.52   & 80.02   & 83.06   & 81.50    \\ \hline
$\mathcal{D}^{\prime}_{\textrm{neg}}$        & 52.78  & 60.96 & 56.58 & 56.24 & 48.43 & 52.04 & 76.45 & 84.29  & 80.18 \\
$\mathcal{D}^{\prime}_{\textrm{pos}}$        & 59.16  & 55.35 & 57.19 & 62.54  & 45.24 & 52.49 & 90.76 & 66.92 & 77.04 \\
$\mathcal{D}$          & 60.63    & 44.65   & 51.42   & 62.72   & 43.45   & 51.34   & 89.44   & 67.57   & 76.98   \\ \hline
\end{tabular}
\end{adjustbox}
\vspace{-5pt}
\caption{Results of only cleaning either negative samples or positive samples.
$\mathcal{D}^{\prime}$ is the cleaned dataset with both and $\mathcal{D}$ is the original distant data.
$\mathcal{D}^{\prime}_{\textrm{pos}}$ and $\mathcal{D}^{\prime}_{\textrm{neg}}$ denote cleaned datasets only consider positive samples and negative samples, respectively.}
\label{tab:more_ablation}
\vspace{-10pt}
\end{table*}

\begin{figure*}[t]
    \centering
    \includegraphics[width=1\textwidth]{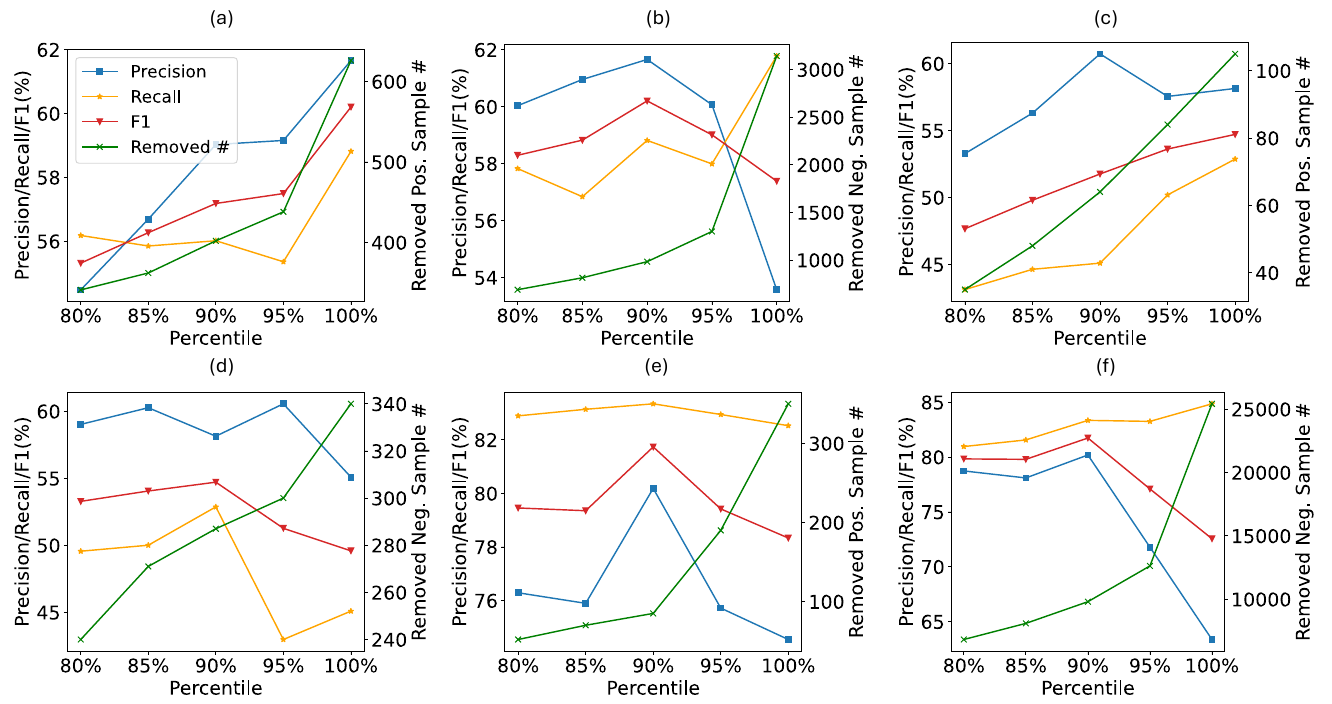}
    \vspace{-15pt}
    \caption{Ablation of varying percentile in threshold samples to compute the thresholds on different datasets. Subfigure (a) and (b) represent the WikiGold dataset, subfigure (c) and (d) represent the WNUT16 dataset, and subfigure (e) and (f) represent the BC5CDR dataset.
    }
    \label{fig:more_ablation}
    \vspace{-10pt}
\end{figure*}

\noindent\textbf{Varying threshold sample percentile.}\quad 
Figure \ref{fig:more_ablation} presents the performance and the number of discarded samples associated with various percentiles on the remaining datasets.
The results shows the $k_{\textrm{pos}}$ and $k_{\textrm{neg}}$ are relative consistent across different datasets.

\section{Additional Analysis}
\subsection{Effectiveness of Cleaning} \label{apx:cleaning_effect}
To further understand the effectiveness of our cleaning method, we using the human annotated CoNLL03 data as ground truth to annotated labels in DS and Cleaned DS $\mathcal{D}^{\prime}_4$, respectively.
We then compute the number true positive labels and false labels (both false negative and false positive) in in DS and Cleaned DS.
We provide the results in Table \ref{tab:clean_results}.
In the Table \ref{tab:clean_results}, the ``\# Pos.'' denotes the correct positive labels for each class. “\# False Neg. + \# False Pos.” are two types of noisy samples, where the ``\# False Pos.'' refers to incorrect labels for each positive class (e.g., a PER span is labeled as LOC span or non-Entity span), ``\# False Neg.'' refers to an entity span missed by distant supervision (e.g., a PER entity is not in the knowledge base, but exists in the text). Considering both is because the DS-NER performance is impacted by two kinds of noise, i.e., false negatives and false positives.

We can find that our methods signficantly reduce the total number of false annotations in original DS data. This is also why our method can improve the performance.
Particularly,
we notice that there is a large number of missing annotations of MISC class, i.e., 2541  false negatives, in the distant supervision step. Thus, the small number of positives and the large number of false negatives cause the model to overfit when trained on the original DS data. 
After cleaning, 
we significantly decrease the number of false annotations, resulting
the performance increases from 29.96 to 73.20 F1 score as shown in Figure \ref{fig:curve}.

\begin{table}[t]
\begin{adjustbox}{width=0.48\textwidth}
\centering
\begin{tabular}{clcccc}
\hline
                            &                            & ORG  & MISC & PER  & LOC  \\ \hline
\multirow{3}{*}{DS}         & \# Pos.                    & 4128 & 786  & 7535 & 5327 \\
                            & \# True Pos.               & 3754 & 786  & 4782 & 5326 \\
                            & \# False Pos. + False Neg. & 2668 & 2541 & 4565 & 1500 \\ \hline
\multirow{3}{*}{Cleaned DS} & \# Pos.                    & 3387 & 709  & 3933 & 5070 \\
                            & \# True Pos.               & 3241 & 709  & 3446 & 5069 \\
                            & \# False Pos. + False Neg. & 293  & 481  & 552  & 388  \\ \hline
\end{tabular}
\end{adjustbox}
\caption{The number of positive labels, the number of true positives, and the corresponding number of false positives and false negatives for each entity type. The results are obtained by comparing the human annotated labels with DS and Cleaned DS $\mathcal{D}^{\prime}_4$, respectively.}
\label{tab:clean_results}
\end{table}

\begin{figure*}[!htbp]
    \centering
    \includegraphics[width=1\textwidth]{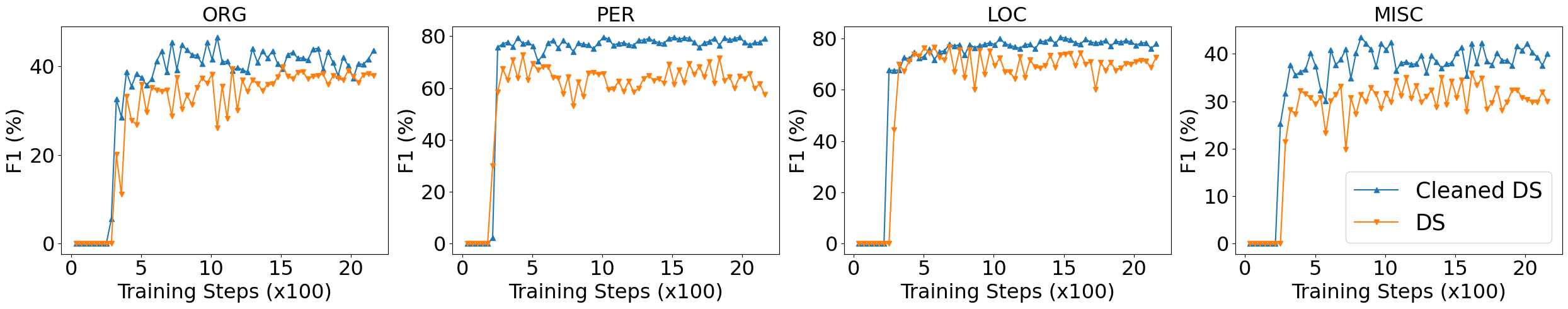}
    \vspace{-15pt}
    \caption{The performance curve for each class of WikiGold when training with the original DS and cleaned DS $\mathcal{D}^{\prime}_4$ and testing on the dev set. ``ORG'', ``PER'', ``LOC'', and ``MISC'' represent the entity types of organization, person, location, and miscellaneous, respectively. 
    }
    \label{fig:wiki_curve}
\end{figure*}

\subsection{Effectiveness of Noise Cleaning on WikiGold} \label{sec:wikigold_cleaning}

We conduct similar noise cleaning analysis on WikiGold, as it has lower KB-Matching performance. 
We using the original DS and cleaned DS $\mathcal{D}^{\prime}_4$ of WikiGold to train the RoBERTa and plot the class-wise performance curve on dev set, as shown in Figure \ref{fig:wiki_curve}.
We find that the model trained on our cleaned data consistently outperforms across classes than the model trained on original DS data, though we used less (but of higher-quality) training data.

Similar to Appendix \ref{apx:cleaning_effect}, we compute the number of true positive labels and false labels (both false negative and false positive) in in DS and Cleaned DS. Table \ref{tab:wiki_clean_results} shows the detailed cleaning effectiveness.
We can find that our method also signficantly reduces the total number of false annotations in original DS data. Thus, the models trained our cleaned dataset achieve better performances.

\begin{table}[H]
\begin{adjustbox}{width=0.48\textwidth}
\centering
\begin{tabular}{clcccc}
\hline
                            &                            & ORG  & MISC & PER  & LOC  \\ \hline
\multirow{3}{*}{DS}         & \# Pos.                    & 715 & 440  & 704 & 421 \\
                            & \# True Pos.               & 240 & 159  & 358 & 335 \\
                            & \# False Pos. + False Neg. & 745 & 529 & 567 & 325 \\ \hline
\multirow{3}{*}{Cleaned DS} & \# Pos.                    & 511 & 382  & 399 & 365 \\
                            & \# True Pos.               & 195 & 143  & 324 & 322 \\
                            & \# False Pos. + False Neg. & 507  & 418  & 176  & 138  \\ \hline
\end{tabular}
\end{adjustbox}
\caption{The number of positive labels, the number of true positives, and the corresponding number of false positives and false negatives for each entity type. The results are obtained by comparing the human annotated labels with DS and Cleaned DS $\mathcal{D}^{\prime}_4$, respectively.}
\label{tab:wiki_clean_results}
\end{table}


\section{Other Training Dynamic Metric} 
\subsection{Definition of Confidence and Variability} \label{apx:conf_define}
The ``Confidence'' score is defined as the mean model probability of the assigned label $y^*$ (potential error) across epochs:

\begin{equation} \label{eq:confidence}
    \hat{\mu}(\mathbf{x},y^*) = \frac{1}{E}\sum_{e=1}^{E}P(\mathbf{x}, y^*)
\end{equation}

Where $P$ denotes the model's probability at the end of $e^{th}$ epoch during training. The ``Variability'' is defined as the standard deviation of $P$ across epochs $E$:
\begin{equation} \label{eq:variability}
    \hat{\sigma} = \sqrt{\frac{\sum_{e=1}^{E} \left( P(y^*| \mathbf{x}) - \hat{\mu}(\mathbf{x},y^*) \right)^2}{E}}
\end{equation}

These two metrics are used to evaluate the characteristics of each individual sample. When using in our DynClean, we directly replace the \ref{eq:AUM} with \ref{eq:confidence} in Algorithm \ref{alg:noise_filter}.

\subsection{Data Map Visulization of AUM and Confidence}\label{apx:datamap}

\begin{figure}[H]
\centering
\includegraphics[width=\linewidth]{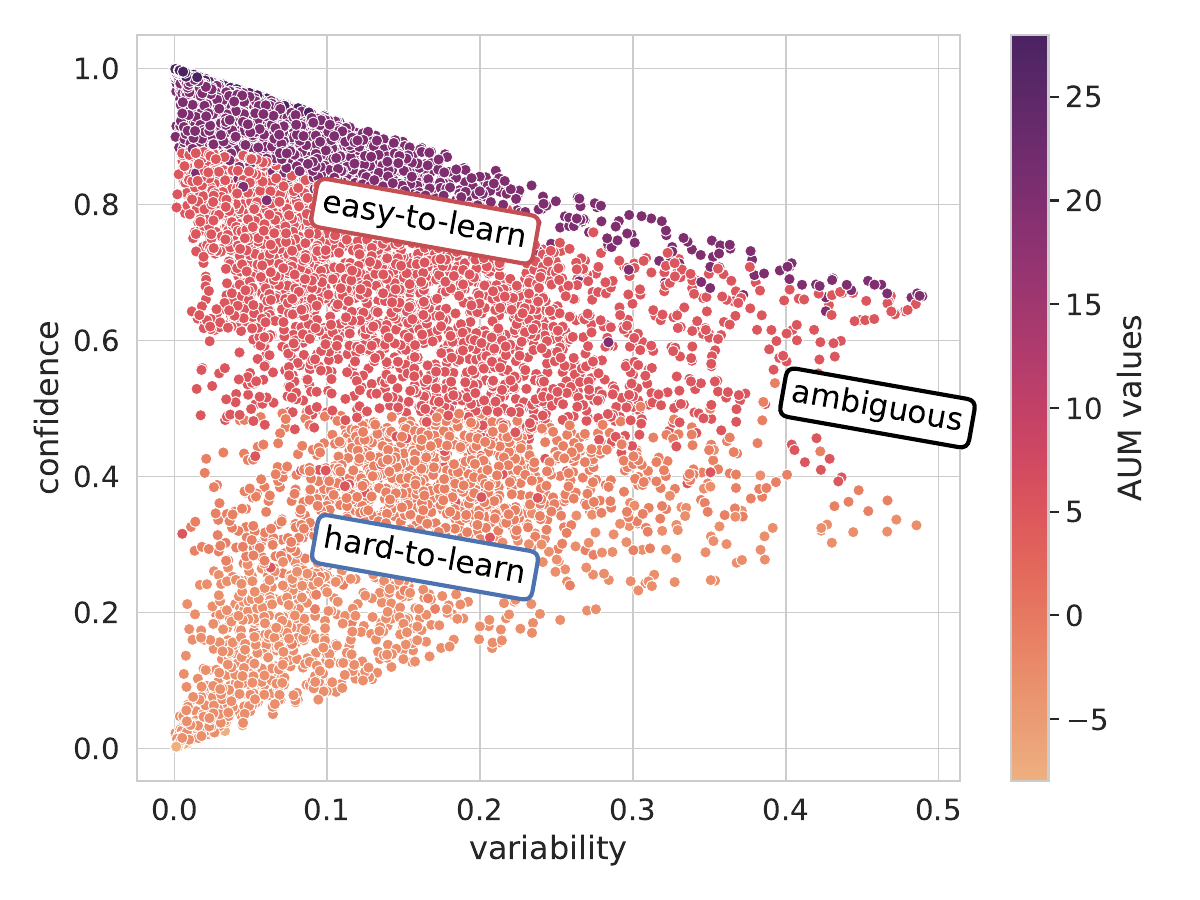}
\caption{Data Map of the positive samples in CoNLL03. We can find that the samples with lower AUM values are mainly in the \textbf{hard-to-learn} region.}
\label{fig:datamap_conll03}
\end{figure}

We visualize the AUM values with their data map as defined by \citep{swayamdipta-etal-2020-dataset} in Figure \ref{fig:datamap_conll03} for positive samples of CoNLL03.
We observe that existing a significant overlap between samples with low AUM values and those with low Confidence values.
This futher shows that the two metrics have very similar effects on identify false samples.

\end{document}